\documentclass{llncs}
\pdfoutput=1

\usepackage[utf8]{inputenc} 
\usepackage[T1]{fontenc}    
\usepackage[hidelinks]{hyperref}       
\usepackage{url}            
\usepackage{booktabs}       
\usepackage{amsfonts}       
\usepackage{nicefrac}       
\usepackage{microtype}      
\usepackage{xcolor}         

\usepackage{colortbl}

\usepackage{booktabs}
\usepackage{amsmath}
\usepackage{hyperref}
\usepackage{graphicx}
\graphicspath{ {./images/} }
\usepackage[final]{pdfpages}

\usepackage{amssymb}
\usepackage{amsfonts}       
\newcommand{\norm}[1]{\left\lVert#1\right\rVert}

\usepackage{booktabs}
\usepackage{subfig}
\usepackage{multicol}
\usepackage{tabularx}
\usepackage{multirow}
\usepackage{xspace}
\usepackage{vector} 
\renewcommand{\vec}[1]{\vect{#1}}
\newcommand{\mat}[1]{\vec{#1}}

\newcommand{\xhdr}[1]{{\noindent\bfseries #1}.}
\usepackage{subfiles}
\usepackage[noabbrev]{cleveref}
\usepackage[backend=bibtex,style=ieee,natbib=true,url=false,doi=false,mincitenames=1,maxnames=99]{biblatex}
\AtBeginBibliography{\small}
\usepackage[mode=buildnew]{standalone}

\RequirePackage{luatex85}

\usepackage{pgfplots}
\pgfplotsset{compat=newest}
\usepgfplotslibrary{groupplots}
\usepgfplotslibrary{polar}
\usepgfplotslibrary{smithchart}
\usepgfplotslibrary{statistics}
\usepgfplotslibrary{dateplot}
\usetikzlibrary{arrows.meta}
\usetikzlibrary{backgrounds}
\usepgfplotslibrary{patchplots}
\usepgfplotslibrary{fillbetween}
\pgfplotsset{%
layers/standard/.define layer set={%
    background,axis background,axis grid,axis ticks,axis lines,axis tick labels,pre main,main,axis descriptions,axis foreground%
}{grid style= {/pgfplots/on layer=axis grid},%
    tick style= {/pgfplots/on layer=axis ticks},%
    axis line style= {/pgfplots/on layer=axis lines},%
    label style= {/pgfplots/on layer=axis descriptions},%
    legend style= {/pgfplots/on layer=axis descriptions},%
    title style= {/pgfplots/on layer=axis descriptions},%
    colorbar style= {/pgfplots/on layer=axis descriptions},%
    ticklabel style= {/pgfplots/on layer=axis tick labels},%
    axis background@ style={/pgfplots/on layer=axis background},%
    3d box foreground style={/pgfplots/on layer=axis foreground},%
    },
}
\pgfplotsset{%
layers/standard/.define layer set={%
    background,axis background,axis grid,axis ticks,axis lines,axis tick labels,pre main,main,axis descriptions,axis foreground%
}{grid style= {/pgfplots/on layer=axis grid},%
    tick style= {/pgfplots/on layer=axis ticks},%
    axis line style= {/pgfplots/on layer=axis lines},%
    label style= {/pgfplots/on layer=axis descriptions},%
    legend style= {/pgfplots/on layer=axis descriptions},%
    title style= {/pgfplots/on layer=axis descriptions},%
    colorbar style= {/pgfplots/on layer=axis descriptions},%
    ticklabel style= {/pgfplots/on layer=axis tick labels},%
    axis background@ style={/pgfplots/on layer=axis background},%
    3d box foreground style={/pgfplots/on layer=axis foreground},%
    },
}

\pgfplotsset{
colormap={plots1}{rgb(0.00000000)=(0.00146200,0.00046600,0.01386600)
rgb(0.00392157)=(0.00226700,0.00127000,0.01857000)
rgb(0.00784314)=(0.00329900,0.00224900,0.02423900)
rgb(0.01176471)=(0.00454700,0.00339200,0.03090900)
rgb(0.01568627)=(0.00600600,0.00469200,0.03855800)
rgb(0.01960784)=(0.00767600,0.00613600,0.04683600)
rgb(0.02352941)=(0.00956100,0.00771300,0.05514300)
rgb(0.02745098)=(0.01166300,0.00941700,0.06346000)
rgb(0.03137255)=(0.01399500,0.01122500,0.07186200)
rgb(0.03529412)=(0.01656100,0.01313600,0.08028200)
rgb(0.03921569)=(0.01937300,0.01513300,0.08876700)
rgb(0.04313725)=(0.02244700,0.01719900,0.09732700)
rgb(0.04705882)=(0.02579300,0.01933100,0.10593000)
rgb(0.05098039)=(0.02943200,0.02150300,0.11462100)
rgb(0.05490196)=(0.03338500,0.02370200,0.12339700)
rgb(0.05882353)=(0.03766800,0.02592100,0.13223200)
rgb(0.06274510)=(0.04225300,0.02813900,0.14114100)
rgb(0.06666667)=(0.04691500,0.03032400,0.15016400)
rgb(0.07058824)=(0.05164400,0.03247400,0.15925400)
rgb(0.07450980)=(0.05644900,0.03456900,0.16841400)
rgb(0.07843137)=(0.06134000,0.03659000,0.17764200)
rgb(0.08235294)=(0.06633100,0.03850400,0.18696200)
rgb(0.08627451)=(0.07142900,0.04029400,0.19635400)
rgb(0.09019608)=(0.07663700,0.04190500,0.20579900)
rgb(0.09411765)=(0.08196200,0.04332800,0.21528900)
rgb(0.09803922)=(0.08741100,0.04455600,0.22481300)
rgb(0.10196078)=(0.09299000,0.04558300,0.23435800)
rgb(0.10588235)=(0.09870200,0.04640200,0.24390400)
rgb(0.10980392)=(0.10455100,0.04700800,0.25343000)
rgb(0.11372549)=(0.11053600,0.04739900,0.26291200)
rgb(0.11764706)=(0.11665600,0.04757400,0.27232100)
rgb(0.12156863)=(0.12290800,0.04753600,0.28162400)
rgb(0.12549020)=(0.12928500,0.04729300,0.29078800)
rgb(0.12941176)=(0.13577800,0.04685600,0.29977600)
rgb(0.13333333)=(0.14237800,0.04624200,0.30855300)
rgb(0.13725490)=(0.14907300,0.04546800,0.31708500)
rgb(0.14117647)=(0.15585000,0.04455900,0.32533800)
rgb(0.14509804)=(0.16268900,0.04355400,0.33327700)
rgb(0.14901961)=(0.16957500,0.04248900,0.34087400)
rgb(0.15294118)=(0.17649300,0.04140200,0.34811100)
rgb(0.15686275)=(0.18342900,0.04032900,0.35497100)
rgb(0.16078431)=(0.19036700,0.03930900,0.36144700)
rgb(0.16470588)=(0.19729700,0.03840000,0.36753500)
rgb(0.16862745)=(0.20420900,0.03763200,0.37323800)
rgb(0.17254902)=(0.21109500,0.03703000,0.37856300)
rgb(0.17647059)=(0.21794900,0.03661500,0.38352200)
rgb(0.18039216)=(0.22476300,0.03640500,0.38812900)
rgb(0.18431373)=(0.23153800,0.03640500,0.39240000)
rgb(0.18823529)=(0.23827300,0.03662100,0.39635300)
rgb(0.19215686)=(0.24496700,0.03705500,0.40000700)
rgb(0.19607843)=(0.25162000,0.03770500,0.40337800)
rgb(0.20000000)=(0.25823400,0.03857100,0.40648500)
rgb(0.20392157)=(0.26481000,0.03964700,0.40934500)
rgb(0.20784314)=(0.27134700,0.04092200,0.41197600)
rgb(0.21176471)=(0.27785000,0.04235300,0.41439200)
rgb(0.21568627)=(0.28432100,0.04393300,0.41660800)
rgb(0.21960784)=(0.29076300,0.04564400,0.41863700)
rgb(0.22352941)=(0.29717800,0.04747000,0.42049100)
rgb(0.22745098)=(0.30356800,0.04939600,0.42218200)
rgb(0.23137255)=(0.30993500,0.05140700,0.42372100)
rgb(0.23529412)=(0.31628200,0.05349000,0.42511600)
rgb(0.23921569)=(0.32261000,0.05563400,0.42637700)
rgb(0.24313725)=(0.32892100,0.05782700,0.42751100)
rgb(0.24705882)=(0.33521700,0.06006000,0.42852400)
rgb(0.25098039)=(0.34150000,0.06232500,0.42942500)
rgb(0.25490196)=(0.34777100,0.06461600,0.43021700)
rgb(0.25882353)=(0.35403200,0.06692500,0.43090600)
rgb(0.26274510)=(0.36028400,0.06924700,0.43149700)
rgb(0.26666667)=(0.36652900,0.07157900,0.43199400)
rgb(0.27058824)=(0.37276800,0.07391500,0.43240000)
rgb(0.27450980)=(0.37900100,0.07625300,0.43271900)
rgb(0.27843137)=(0.38522800,0.07859100,0.43295500)
rgb(0.28235294)=(0.39145300,0.08092700,0.43310900)
rgb(0.28627451)=(0.39767400,0.08325700,0.43318300)
rgb(0.29019608)=(0.40389400,0.08558000,0.43317900)
rgb(0.29411765)=(0.41011300,0.08789600,0.43309800)
rgb(0.29803922)=(0.41633100,0.09020300,0.43294300)
rgb(0.30196078)=(0.42254900,0.09250100,0.43271400)
rgb(0.30588235)=(0.42876800,0.09479000,0.43241200)
rgb(0.30980392)=(0.43498700,0.09706900,0.43203900)
rgb(0.31372549)=(0.44120700,0.09933800,0.43159400)
rgb(0.31764706)=(0.44742800,0.10159700,0.43108000)
rgb(0.32156863)=(0.45365100,0.10384800,0.43049800)
rgb(0.32549020)=(0.45987500,0.10608900,0.42984600)
rgb(0.32941176)=(0.46610000,0.10832200,0.42912500)
rgb(0.33333333)=(0.47232800,0.11054700,0.42833400)
rgb(0.33725490)=(0.47855800,0.11276400,0.42747500)
rgb(0.34117647)=(0.48478900,0.11497400,0.42654800)
rgb(0.34509804)=(0.49102200,0.11717900,0.42555200)
rgb(0.34901961)=(0.49725700,0.11937900,0.42448800)
rgb(0.35294118)=(0.50349300,0.12157500,0.42335600)
rgb(0.35686275)=(0.50973000,0.12376900,0.42215600)
rgb(0.36078431)=(0.51596700,0.12596000,0.42088700)
rgb(0.36470588)=(0.52220600,0.12815000,0.41954900)
rgb(0.36862745)=(0.52844400,0.13034100,0.41814200)
rgb(0.37254902)=(0.53468300,0.13253400,0.41666700)
rgb(0.37647059)=(0.54092000,0.13472900,0.41512300)
rgb(0.38039216)=(0.54715700,0.13692900,0.41351100)
rgb(0.38431373)=(0.55339200,0.13913400,0.41182900)
rgb(0.38823529)=(0.55962400,0.14134600,0.41007800)
rgb(0.39215686)=(0.56585400,0.14356700,0.40825800)
rgb(0.39607843)=(0.57208100,0.14579700,0.40636900)
rgb(0.40000000)=(0.57830400,0.14803900,0.40441100)
rgb(0.40392157)=(0.58452100,0.15029400,0.40238500)
rgb(0.40784314)=(0.59073400,0.15256300,0.40029000)
rgb(0.41176471)=(0.59694000,0.15484800,0.39812500)
rgb(0.41568627)=(0.60313900,0.15715100,0.39589100)
rgb(0.41960784)=(0.60933000,0.15947400,0.39358900)
rgb(0.42352941)=(0.61551300,0.16181700,0.39121900)
rgb(0.42745098)=(0.62168500,0.16418400,0.38878100)
rgb(0.43137255)=(0.62784700,0.16657500,0.38627600)
rgb(0.43529412)=(0.63399800,0.16899200,0.38370400)
rgb(0.43921569)=(0.64013500,0.17143800,0.38106500)
rgb(0.44313725)=(0.64626000,0.17391400,0.37835900)
rgb(0.44705882)=(0.65236900,0.17642100,0.37558600)
rgb(0.45098039)=(0.65846300,0.17896200,0.37274800)
rgb(0.45490196)=(0.66454000,0.18153900,0.36984600)
rgb(0.45882353)=(0.67059900,0.18415300,0.36687900)
rgb(0.46274510)=(0.67663800,0.18680700,0.36384900)
rgb(0.46666667)=(0.68265600,0.18950100,0.36075700)
rgb(0.47058824)=(0.68865300,0.19223900,0.35760300)
rgb(0.47450980)=(0.69462700,0.19502100,0.35438800)
rgb(0.47843137)=(0.70057600,0.19785100,0.35111300)
rgb(0.48235294)=(0.70650000,0.20072800,0.34777700)
rgb(0.48627451)=(0.71239600,0.20365600,0.34438300)
rgb(0.49019608)=(0.71826400,0.20663600,0.34093100)
rgb(0.49411765)=(0.72410300,0.20967000,0.33742400)
rgb(0.49803922)=(0.72990900,0.21275900,0.33386100)
rgb(0.50196078)=(0.73568300,0.21590600,0.33024500)
rgb(0.50588235)=(0.74142300,0.21911200,0.32657600)
rgb(0.50980392)=(0.74712700,0.22237800,0.32285600)
rgb(0.51372549)=(0.75279400,0.22570600,0.31908500)
rgb(0.51764706)=(0.75842200,0.22909700,0.31526600)
rgb(0.52156863)=(0.76401000,0.23255400,0.31139900)
rgb(0.52549020)=(0.76955600,0.23607700,0.30748500)
rgb(0.52941176)=(0.77505900,0.23966700,0.30352600)
rgb(0.53333333)=(0.78051700,0.24332700,0.29952300)
rgb(0.53725490)=(0.78592900,0.24705600,0.29547700)
rgb(0.54117647)=(0.79129300,0.25085600,0.29139000)
rgb(0.54509804)=(0.79660700,0.25472800,0.28726400)
rgb(0.54901961)=(0.80187100,0.25867400,0.28309900)
rgb(0.55294118)=(0.80708200,0.26269200,0.27889800)
rgb(0.55686275)=(0.81223900,0.26678600,0.27466100)
rgb(0.56078431)=(0.81734100,0.27095400,0.27039000)
rgb(0.56470588)=(0.82238600,0.27519700,0.26608500)
rgb(0.56862745)=(0.82737200,0.27951700,0.26175000)
rgb(0.57254902)=(0.83229900,0.28391300,0.25738300)
rgb(0.57647059)=(0.83716500,0.28838500,0.25298800)
rgb(0.58039216)=(0.84196900,0.29293300,0.24856400)
rgb(0.58431373)=(0.84670900,0.29755900,0.24411300)
rgb(0.58823529)=(0.85138400,0.30226000,0.23963600)
rgb(0.59215686)=(0.85599200,0.30703800,0.23513300)
rgb(0.59607843)=(0.86053300,0.31189200,0.23060600)
rgb(0.60000000)=(0.86500600,0.31682200,0.22605500)
rgb(0.60392157)=(0.86940900,0.32182700,0.22148200)
rgb(0.60784314)=(0.87374100,0.32690600,0.21688600)
rgb(0.61176471)=(0.87800100,0.33206000,0.21226800)
rgb(0.61568627)=(0.88218800,0.33728700,0.20762800)
rgb(0.61960784)=(0.88630200,0.34258600,0.20296800)
rgb(0.62352941)=(0.89034100,0.34795700,0.19828600)
rgb(0.62745098)=(0.89430500,0.35339900,0.19358400)
rgb(0.63137255)=(0.89819200,0.35891100,0.18886000)
rgb(0.63529412)=(0.90200300,0.36449200,0.18411600)
rgb(0.63921569)=(0.90573500,0.37014000,0.17935000)
rgb(0.64313725)=(0.90939000,0.37585600,0.17456300)
rgb(0.64705882)=(0.91296600,0.38163600,0.16975500)
rgb(0.65098039)=(0.91646200,0.38748100,0.16492400)
rgb(0.65490196)=(0.91987900,0.39338900,0.16007000)
rgb(0.65882353)=(0.92321500,0.39935900,0.15519300)
rgb(0.66274510)=(0.92647000,0.40538900,0.15029200)
rgb(0.66666667)=(0.92964400,0.41147900,0.14536700)
rgb(0.67058824)=(0.93273700,0.41762700,0.14041700)
rgb(0.67450980)=(0.93574700,0.42383100,0.13544000)
rgb(0.67843137)=(0.93867500,0.43009100,0.13043800)
rgb(0.68235294)=(0.94152100,0.43640500,0.12540900)
rgb(0.68627451)=(0.94428500,0.44277200,0.12035400)
rgb(0.69019608)=(0.94696500,0.44919100,0.11527200)
rgb(0.69411765)=(0.94956200,0.45566000,0.11016400)
rgb(0.69803922)=(0.95207500,0.46217800,0.10503100)
rgb(0.70196078)=(0.95450600,0.46874400,0.09987400)
rgb(0.70588235)=(0.95685200,0.47535600,0.09469500)
rgb(0.70980392)=(0.95911400,0.48201400,0.08949900)
rgb(0.71372549)=(0.96129300,0.48871600,0.08428900)
rgb(0.71764706)=(0.96338700,0.49546200,0.07907300)
rgb(0.72156863)=(0.96539700,0.50224900,0.07385900)
rgb(0.72549020)=(0.96732200,0.50907800,0.06865900)
rgb(0.72941176)=(0.96916300,0.51594600,0.06348800)
rgb(0.73333333)=(0.97091900,0.52285300,0.05836700)
rgb(0.73725490)=(0.97259000,0.52979800,0.05332400)
rgb(0.74117647)=(0.97417600,0.53678000,0.04839200)
rgb(0.74509804)=(0.97567700,0.54379800,0.04361800)
rgb(0.74901961)=(0.97709200,0.55085000,0.03905000)
rgb(0.75294118)=(0.97842200,0.55793700,0.03493100)
rgb(0.75686275)=(0.97966600,0.56505700,0.03140900)
rgb(0.76078431)=(0.98082400,0.57220900,0.02850800)
rgb(0.76470588)=(0.98189500,0.57939200,0.02625000)
rgb(0.76862745)=(0.98288100,0.58660600,0.02466100)
rgb(0.77254902)=(0.98377900,0.59384900,0.02377000)
rgb(0.77647059)=(0.98459100,0.60112200,0.02360600)
rgb(0.78039216)=(0.98531500,0.60842200,0.02420200)
rgb(0.78431373)=(0.98595200,0.61575000,0.02559200)
rgb(0.78823529)=(0.98650200,0.62310500,0.02781400)
rgb(0.79215686)=(0.98696400,0.63048500,0.03090800)
rgb(0.79607843)=(0.98733700,0.63789000,0.03491600)
rgb(0.80000000)=(0.98762200,0.64532000,0.03988600)
rgb(0.80392157)=(0.98781900,0.65277300,0.04558100)
rgb(0.80784314)=(0.98792600,0.66025000,0.05175000)
rgb(0.81176471)=(0.98794500,0.66774800,0.05832900)
rgb(0.81568627)=(0.98787400,0.67526700,0.06525700)
rgb(0.81960784)=(0.98771400,0.68280700,0.07248900)
rgb(0.82352941)=(0.98746400,0.69036600,0.07999000)
rgb(0.82745098)=(0.98712400,0.69794400,0.08773100)
rgb(0.83137255)=(0.98669400,0.70554000,0.09569400)
rgb(0.83529412)=(0.98617500,0.71315300,0.10386300)
rgb(0.83921569)=(0.98556600,0.72078200,0.11222900)
rgb(0.84313725)=(0.98486500,0.72842700,0.12078500)
rgb(0.84705882)=(0.98407500,0.73608700,0.12952700)
rgb(0.85098039)=(0.98319600,0.74375800,0.13845300)
rgb(0.85490196)=(0.98222800,0.75144200,0.14756500)
rgb(0.85882353)=(0.98117300,0.75913500,0.15686300)
rgb(0.86274510)=(0.98003200,0.76683700,0.16635300)
rgb(0.86666667)=(0.97880600,0.77454500,0.17603700)
rgb(0.87058824)=(0.97749700,0.78225800,0.18592300)
rgb(0.87450980)=(0.97610800,0.78997400,0.19601800)
rgb(0.87843137)=(0.97463800,0.79769200,0.20633200)
rgb(0.88235294)=(0.97308800,0.80540900,0.21687700)
rgb(0.88627451)=(0.97146800,0.81312200,0.22765800)
rgb(0.89019608)=(0.96978300,0.82082500,0.23868600)
rgb(0.89411765)=(0.96804100,0.82851500,0.24997200)
rgb(0.89803922)=(0.96624300,0.83619100,0.26153400)
rgb(0.90196078)=(0.96439400,0.84384800,0.27339100)
rgb(0.90588235)=(0.96251700,0.85147600,0.28554600)
rgb(0.90980392)=(0.96062600,0.85906900,0.29801000)
rgb(0.91372549)=(0.95872000,0.86662400,0.31082000)
rgb(0.91764706)=(0.95683400,0.87412900,0.32397400)
rgb(0.92156863)=(0.95499700,0.88156900,0.33747500)
rgb(0.92549020)=(0.95321500,0.88894200,0.35136900)
rgb(0.92941176)=(0.95154600,0.89622600,0.36562700)
rgb(0.93333333)=(0.95001800,0.90340900,0.38027100)
rgb(0.93725490)=(0.94868300,0.91047300,0.39528900)
rgb(0.94117647)=(0.94759400,0.91739900,0.41066500)
rgb(0.94509804)=(0.94680900,0.92416800,0.42637300)
rgb(0.94901961)=(0.94639200,0.93076100,0.44236700)
rgb(0.95294118)=(0.94640300,0.93715900,0.45859200)
rgb(0.95686275)=(0.94690300,0.94334800,0.47497000)
rgb(0.96078431)=(0.94793700,0.94931800,0.49142600)
rgb(0.96470588)=(0.94954500,0.95506300,0.50786000)
rgb(0.96862745)=(0.95174000,0.96058700,0.52420300)
rgb(0.97254902)=(0.95452900,0.96589600,0.54036100)
rgb(0.97647059)=(0.95789600,0.97100300,0.55627500)
rgb(0.98039216)=(0.96181200,0.97592400,0.57192500)
rgb(0.98431373)=(0.96624900,0.98067800,0.58720600)
rgb(0.98823529)=(0.97116200,0.98528200,0.60215400)
rgb(0.99215686)=(0.97651100,0.98975300,0.61676000)
rgb(0.99607843)=(0.98225700,0.99410900,0.63101700)
rgb(1.00000000)=(0.98836200,0.99836400,0.64492400)},
}
\pgfplotsset{
colormap={plots1}{rgb(0.00000000)=(0.00146200,0.00046600,0.01386600)
rgb(0.00392157)=(0.00226700,0.00127000,0.01857000)
rgb(0.00784314)=(0.00329900,0.00224900,0.02423900)
rgb(0.01176471)=(0.00454700,0.00339200,0.03090900)
rgb(0.01568627)=(0.00600600,0.00469200,0.03855800)
rgb(0.01960784)=(0.00767600,0.00613600,0.04683600)
rgb(0.02352941)=(0.00956100,0.00771300,0.05514300)
rgb(0.02745098)=(0.01166300,0.00941700,0.06346000)
rgb(0.03137255)=(0.01399500,0.01122500,0.07186200)
rgb(0.03529412)=(0.01656100,0.01313600,0.08028200)
rgb(0.03921569)=(0.01937300,0.01513300,0.08876700)
rgb(0.04313725)=(0.02244700,0.01719900,0.09732700)
rgb(0.04705882)=(0.02579300,0.01933100,0.10593000)
rgb(0.05098039)=(0.02943200,0.02150300,0.11462100)
rgb(0.05490196)=(0.03338500,0.02370200,0.12339700)
rgb(0.05882353)=(0.03766800,0.02592100,0.13223200)
rgb(0.06274510)=(0.04225300,0.02813900,0.14114100)
rgb(0.06666667)=(0.04691500,0.03032400,0.15016400)
rgb(0.07058824)=(0.05164400,0.03247400,0.15925400)
rgb(0.07450980)=(0.05644900,0.03456900,0.16841400)
rgb(0.07843137)=(0.06134000,0.03659000,0.17764200)
rgb(0.08235294)=(0.06633100,0.03850400,0.18696200)
rgb(0.08627451)=(0.07142900,0.04029400,0.19635400)
rgb(0.09019608)=(0.07663700,0.04190500,0.20579900)
rgb(0.09411765)=(0.08196200,0.04332800,0.21528900)
rgb(0.09803922)=(0.08741100,0.04455600,0.22481300)
rgb(0.10196078)=(0.09299000,0.04558300,0.23435800)
rgb(0.10588235)=(0.09870200,0.04640200,0.24390400)
rgb(0.10980392)=(0.10455100,0.04700800,0.25343000)
rgb(0.11372549)=(0.11053600,0.04739900,0.26291200)
rgb(0.11764706)=(0.11665600,0.04757400,0.27232100)
rgb(0.12156863)=(0.12290800,0.04753600,0.28162400)
rgb(0.12549020)=(0.12928500,0.04729300,0.29078800)
rgb(0.12941176)=(0.13577800,0.04685600,0.29977600)
rgb(0.13333333)=(0.14237800,0.04624200,0.30855300)
rgb(0.13725490)=(0.14907300,0.04546800,0.31708500)
rgb(0.14117647)=(0.15585000,0.04455900,0.32533800)
rgb(0.14509804)=(0.16268900,0.04355400,0.33327700)
rgb(0.14901961)=(0.16957500,0.04248900,0.34087400)
rgb(0.15294118)=(0.17649300,0.04140200,0.34811100)
rgb(0.15686275)=(0.18342900,0.04032900,0.35497100)
rgb(0.16078431)=(0.19036700,0.03930900,0.36144700)
rgb(0.16470588)=(0.19729700,0.03840000,0.36753500)
rgb(0.16862745)=(0.20420900,0.03763200,0.37323800)
rgb(0.17254902)=(0.21109500,0.03703000,0.37856300)
rgb(0.17647059)=(0.21794900,0.03661500,0.38352200)
rgb(0.18039216)=(0.22476300,0.03640500,0.38812900)
rgb(0.18431373)=(0.23153800,0.03640500,0.39240000)
rgb(0.18823529)=(0.23827300,0.03662100,0.39635300)
rgb(0.19215686)=(0.24496700,0.03705500,0.40000700)
rgb(0.19607843)=(0.25162000,0.03770500,0.40337800)
rgb(0.20000000)=(0.25823400,0.03857100,0.40648500)
rgb(0.20392157)=(0.26481000,0.03964700,0.40934500)
rgb(0.20784314)=(0.27134700,0.04092200,0.41197600)
rgb(0.21176471)=(0.27785000,0.04235300,0.41439200)
rgb(0.21568627)=(0.28432100,0.04393300,0.41660800)
rgb(0.21960784)=(0.29076300,0.04564400,0.41863700)
rgb(0.22352941)=(0.29717800,0.04747000,0.42049100)
rgb(0.22745098)=(0.30356800,0.04939600,0.42218200)
rgb(0.23137255)=(0.30993500,0.05140700,0.42372100)
rgb(0.23529412)=(0.31628200,0.05349000,0.42511600)
rgb(0.23921569)=(0.32261000,0.05563400,0.42637700)
rgb(0.24313725)=(0.32892100,0.05782700,0.42751100)
rgb(0.24705882)=(0.33521700,0.06006000,0.42852400)
rgb(0.25098039)=(0.34150000,0.06232500,0.42942500)
rgb(0.25490196)=(0.34777100,0.06461600,0.43021700)
rgb(0.25882353)=(0.35403200,0.06692500,0.43090600)
rgb(0.26274510)=(0.36028400,0.06924700,0.43149700)
rgb(0.26666667)=(0.36652900,0.07157900,0.43199400)
rgb(0.27058824)=(0.37276800,0.07391500,0.43240000)
rgb(0.27450980)=(0.37900100,0.07625300,0.43271900)
rgb(0.27843137)=(0.38522800,0.07859100,0.43295500)
rgb(0.28235294)=(0.39145300,0.08092700,0.43310900)
rgb(0.28627451)=(0.39767400,0.08325700,0.43318300)
rgb(0.29019608)=(0.40389400,0.08558000,0.43317900)
rgb(0.29411765)=(0.41011300,0.08789600,0.43309800)
rgb(0.29803922)=(0.41633100,0.09020300,0.43294300)
rgb(0.30196078)=(0.42254900,0.09250100,0.43271400)
rgb(0.30588235)=(0.42876800,0.09479000,0.43241200)
rgb(0.30980392)=(0.43498700,0.09706900,0.43203900)
rgb(0.31372549)=(0.44120700,0.09933800,0.43159400)
rgb(0.31764706)=(0.44742800,0.10159700,0.43108000)
rgb(0.32156863)=(0.45365100,0.10384800,0.43049800)
rgb(0.32549020)=(0.45987500,0.10608900,0.42984600)
rgb(0.32941176)=(0.46610000,0.10832200,0.42912500)
rgb(0.33333333)=(0.47232800,0.11054700,0.42833400)
rgb(0.33725490)=(0.47855800,0.11276400,0.42747500)
rgb(0.34117647)=(0.48478900,0.11497400,0.42654800)
rgb(0.34509804)=(0.49102200,0.11717900,0.42555200)
rgb(0.34901961)=(0.49725700,0.11937900,0.42448800)
rgb(0.35294118)=(0.50349300,0.12157500,0.42335600)
rgb(0.35686275)=(0.50973000,0.12376900,0.42215600)
rgb(0.36078431)=(0.51596700,0.12596000,0.42088700)
rgb(0.36470588)=(0.52220600,0.12815000,0.41954900)
rgb(0.36862745)=(0.52844400,0.13034100,0.41814200)
rgb(0.37254902)=(0.53468300,0.13253400,0.41666700)
rgb(0.37647059)=(0.54092000,0.13472900,0.41512300)
rgb(0.38039216)=(0.54715700,0.13692900,0.41351100)
rgb(0.38431373)=(0.55339200,0.13913400,0.41182900)
rgb(0.38823529)=(0.55962400,0.14134600,0.41007800)
rgb(0.39215686)=(0.56585400,0.14356700,0.40825800)
rgb(0.39607843)=(0.57208100,0.14579700,0.40636900)
rgb(0.40000000)=(0.57830400,0.14803900,0.40441100)
rgb(0.40392157)=(0.58452100,0.15029400,0.40238500)
rgb(0.40784314)=(0.59073400,0.15256300,0.40029000)
rgb(0.41176471)=(0.59694000,0.15484800,0.39812500)
rgb(0.41568627)=(0.60313900,0.15715100,0.39589100)
rgb(0.41960784)=(0.60933000,0.15947400,0.39358900)
rgb(0.42352941)=(0.61551300,0.16181700,0.39121900)
rgb(0.42745098)=(0.62168500,0.16418400,0.38878100)
rgb(0.43137255)=(0.62784700,0.16657500,0.38627600)
rgb(0.43529412)=(0.63399800,0.16899200,0.38370400)
rgb(0.43921569)=(0.64013500,0.17143800,0.38106500)
rgb(0.44313725)=(0.64626000,0.17391400,0.37835900)
rgb(0.44705882)=(0.65236900,0.17642100,0.37558600)
rgb(0.45098039)=(0.65846300,0.17896200,0.37274800)
rgb(0.45490196)=(0.66454000,0.18153900,0.36984600)
rgb(0.45882353)=(0.67059900,0.18415300,0.36687900)
rgb(0.46274510)=(0.67663800,0.18680700,0.36384900)
rgb(0.46666667)=(0.68265600,0.18950100,0.36075700)
rgb(0.47058824)=(0.68865300,0.19223900,0.35760300)
rgb(0.47450980)=(0.69462700,0.19502100,0.35438800)
rgb(0.47843137)=(0.70057600,0.19785100,0.35111300)
rgb(0.48235294)=(0.70650000,0.20072800,0.34777700)
rgb(0.48627451)=(0.71239600,0.20365600,0.34438300)
rgb(0.49019608)=(0.71826400,0.20663600,0.34093100)
rgb(0.49411765)=(0.72410300,0.20967000,0.33742400)
rgb(0.49803922)=(0.72990900,0.21275900,0.33386100)
rgb(0.50196078)=(0.73568300,0.21590600,0.33024500)
rgb(0.50588235)=(0.74142300,0.21911200,0.32657600)
rgb(0.50980392)=(0.74712700,0.22237800,0.32285600)
rgb(0.51372549)=(0.75279400,0.22570600,0.31908500)
rgb(0.51764706)=(0.75842200,0.22909700,0.31526600)
rgb(0.52156863)=(0.76401000,0.23255400,0.31139900)
rgb(0.52549020)=(0.76955600,0.23607700,0.30748500)
rgb(0.52941176)=(0.77505900,0.23966700,0.30352600)
rgb(0.53333333)=(0.78051700,0.24332700,0.29952300)
rgb(0.53725490)=(0.78592900,0.24705600,0.29547700)
rgb(0.54117647)=(0.79129300,0.25085600,0.29139000)
rgb(0.54509804)=(0.79660700,0.25472800,0.28726400)
rgb(0.54901961)=(0.80187100,0.25867400,0.28309900)
rgb(0.55294118)=(0.80708200,0.26269200,0.27889800)
rgb(0.55686275)=(0.81223900,0.26678600,0.27466100)
rgb(0.56078431)=(0.81734100,0.27095400,0.27039000)
rgb(0.56470588)=(0.82238600,0.27519700,0.26608500)
rgb(0.56862745)=(0.82737200,0.27951700,0.26175000)
rgb(0.57254902)=(0.83229900,0.28391300,0.25738300)
rgb(0.57647059)=(0.83716500,0.28838500,0.25298800)
rgb(0.58039216)=(0.84196900,0.29293300,0.24856400)
rgb(0.58431373)=(0.84670900,0.29755900,0.24411300)
rgb(0.58823529)=(0.85138400,0.30226000,0.23963600)
rgb(0.59215686)=(0.85599200,0.30703800,0.23513300)
rgb(0.59607843)=(0.86053300,0.31189200,0.23060600)
rgb(0.60000000)=(0.86500600,0.31682200,0.22605500)
rgb(0.60392157)=(0.86940900,0.32182700,0.22148200)
rgb(0.60784314)=(0.87374100,0.32690600,0.21688600)
rgb(0.61176471)=(0.87800100,0.33206000,0.21226800)
rgb(0.61568627)=(0.88218800,0.33728700,0.20762800)
rgb(0.61960784)=(0.88630200,0.34258600,0.20296800)
rgb(0.62352941)=(0.89034100,0.34795700,0.19828600)
rgb(0.62745098)=(0.89430500,0.35339900,0.19358400)
rgb(0.63137255)=(0.89819200,0.35891100,0.18886000)
rgb(0.63529412)=(0.90200300,0.36449200,0.18411600)
rgb(0.63921569)=(0.90573500,0.37014000,0.17935000)
rgb(0.64313725)=(0.90939000,0.37585600,0.17456300)
rgb(0.64705882)=(0.91296600,0.38163600,0.16975500)
rgb(0.65098039)=(0.91646200,0.38748100,0.16492400)
rgb(0.65490196)=(0.91987900,0.39338900,0.16007000)
rgb(0.65882353)=(0.92321500,0.39935900,0.15519300)
rgb(0.66274510)=(0.92647000,0.40538900,0.15029200)
rgb(0.66666667)=(0.92964400,0.41147900,0.14536700)
rgb(0.67058824)=(0.93273700,0.41762700,0.14041700)
rgb(0.67450980)=(0.93574700,0.42383100,0.13544000)
rgb(0.67843137)=(0.93867500,0.43009100,0.13043800)
rgb(0.68235294)=(0.94152100,0.43640500,0.12540900)
rgb(0.68627451)=(0.94428500,0.44277200,0.12035400)
rgb(0.69019608)=(0.94696500,0.44919100,0.11527200)
rgb(0.69411765)=(0.94956200,0.45566000,0.11016400)
rgb(0.69803922)=(0.95207500,0.46217800,0.10503100)
rgb(0.70196078)=(0.95450600,0.46874400,0.09987400)
rgb(0.70588235)=(0.95685200,0.47535600,0.09469500)
rgb(0.70980392)=(0.95911400,0.48201400,0.08949900)
rgb(0.71372549)=(0.96129300,0.48871600,0.08428900)
rgb(0.71764706)=(0.96338700,0.49546200,0.07907300)
rgb(0.72156863)=(0.96539700,0.50224900,0.07385900)
rgb(0.72549020)=(0.96732200,0.50907800,0.06865900)
rgb(0.72941176)=(0.96916300,0.51594600,0.06348800)
rgb(0.73333333)=(0.97091900,0.52285300,0.05836700)
rgb(0.73725490)=(0.97259000,0.52979800,0.05332400)
rgb(0.74117647)=(0.97417600,0.53678000,0.04839200)
rgb(0.74509804)=(0.97567700,0.54379800,0.04361800)
rgb(0.74901961)=(0.97709200,0.55085000,0.03905000)
rgb(0.75294118)=(0.97842200,0.55793700,0.03493100)
rgb(0.75686275)=(0.97966600,0.56505700,0.03140900)
rgb(0.76078431)=(0.98082400,0.57220900,0.02850800)
rgb(0.76470588)=(0.98189500,0.57939200,0.02625000)
rgb(0.76862745)=(0.98288100,0.58660600,0.02466100)
rgb(0.77254902)=(0.98377900,0.59384900,0.02377000)
rgb(0.77647059)=(0.98459100,0.60112200,0.02360600)
rgb(0.78039216)=(0.98531500,0.60842200,0.02420200)
rgb(0.78431373)=(0.98595200,0.61575000,0.02559200)
rgb(0.78823529)=(0.98650200,0.62310500,0.02781400)
rgb(0.79215686)=(0.98696400,0.63048500,0.03090800)
rgb(0.79607843)=(0.98733700,0.63789000,0.03491600)
rgb(0.80000000)=(0.98762200,0.64532000,0.03988600)
rgb(0.80392157)=(0.98781900,0.65277300,0.04558100)
rgb(0.80784314)=(0.98792600,0.66025000,0.05175000)
rgb(0.81176471)=(0.98794500,0.66774800,0.05832900)
rgb(0.81568627)=(0.98787400,0.67526700,0.06525700)
rgb(0.81960784)=(0.98771400,0.68280700,0.07248900)
rgb(0.82352941)=(0.98746400,0.69036600,0.07999000)
rgb(0.82745098)=(0.98712400,0.69794400,0.08773100)
rgb(0.83137255)=(0.98669400,0.70554000,0.09569400)
rgb(0.83529412)=(0.98617500,0.71315300,0.10386300)
rgb(0.83921569)=(0.98556600,0.72078200,0.11222900)
rgb(0.84313725)=(0.98486500,0.72842700,0.12078500)
rgb(0.84705882)=(0.98407500,0.73608700,0.12952700)
rgb(0.85098039)=(0.98319600,0.74375800,0.13845300)
rgb(0.85490196)=(0.98222800,0.75144200,0.14756500)
rgb(0.85882353)=(0.98117300,0.75913500,0.15686300)
rgb(0.86274510)=(0.98003200,0.76683700,0.16635300)
rgb(0.86666667)=(0.97880600,0.77454500,0.17603700)
rgb(0.87058824)=(0.97749700,0.78225800,0.18592300)
rgb(0.87450980)=(0.97610800,0.78997400,0.19601800)
rgb(0.87843137)=(0.97463800,0.79769200,0.20633200)
rgb(0.88235294)=(0.97308800,0.80540900,0.21687700)
rgb(0.88627451)=(0.97146800,0.81312200,0.22765800)
rgb(0.89019608)=(0.96978300,0.82082500,0.23868600)
rgb(0.89411765)=(0.96804100,0.82851500,0.24997200)
rgb(0.89803922)=(0.96624300,0.83619100,0.26153400)
rgb(0.90196078)=(0.96439400,0.84384800,0.27339100)
rgb(0.90588235)=(0.96251700,0.85147600,0.28554600)
rgb(0.90980392)=(0.96062600,0.85906900,0.29801000)
rgb(0.91372549)=(0.95872000,0.86662400,0.31082000)
rgb(0.91764706)=(0.95683400,0.87412900,0.32397400)
rgb(0.92156863)=(0.95499700,0.88156900,0.33747500)
rgb(0.92549020)=(0.95321500,0.88894200,0.35136900)
rgb(0.92941176)=(0.95154600,0.89622600,0.36562700)
rgb(0.93333333)=(0.95001800,0.90340900,0.38027100)
rgb(0.93725490)=(0.94868300,0.91047300,0.39528900)
rgb(0.94117647)=(0.94759400,0.91739900,0.41066500)
rgb(0.94509804)=(0.94680900,0.92416800,0.42637300)
rgb(0.94901961)=(0.94639200,0.93076100,0.44236700)
rgb(0.95294118)=(0.94640300,0.93715900,0.45859200)
rgb(0.95686275)=(0.94690300,0.94334800,0.47497000)
rgb(0.96078431)=(0.94793700,0.94931800,0.49142600)
rgb(0.96470588)=(0.94954500,0.95506300,0.50786000)
rgb(0.96862745)=(0.95174000,0.96058700,0.52420300)
rgb(0.97254902)=(0.95452900,0.96589600,0.54036100)
rgb(0.97647059)=(0.95789600,0.97100300,0.55627500)
rgb(0.98039216)=(0.96181200,0.97592400,0.57192500)
rgb(0.98431373)=(0.96624900,0.98067800,0.58720600)
rgb(0.98823529)=(0.97116200,0.98528200,0.60215400)
rgb(0.99215686)=(0.97651100,0.98975300,0.61676000)
rgb(0.99607843)=(0.98225700,0.99410900,0.63101700)
rgb(1.00000000)=(0.98836200,0.99836400,0.64492400)},
}

\pgfplotsset{every axis legend/.append style={%
cells={anchor=west}}
}

\pgfplotsset{every axis/.append style={
                    title style={font=\small},
                    tick label style={font=\footnotesize}  
                    }}
\pgfplotsset{every axis label/.style={font=\small}} 
\usetikzlibrary{arrows}
\tikzset{>=stealth'}
\usepackage{algorithm}
\usepackage[noend]{algpseudocode}

\usepackage{siunitx}
\newif\ifanonymize
\anonymizefalse
\newif\ifappendix
\appendixfalse


\title{ZoPE: A Fast Optimizer for ReLU Networks \\ with Low-Dimensional Inputs}

\begin{document}

\ifappendix 

\appendix


\else
%
%
\title{ZoPE: A Fast Optimizer for ReLU Networks \\ with Low-Dimensional Inputs}
%
%
\author{Christopher A. Strong\inst{1} \and
Sydney M. Katz \inst{2} \and \\
Anthony L. Corso \inst{2} \and Mykel J. Kochenderfer \inst{2}}
\authorrunning{C. Strong et al.}
%
\institute{Department of Electrical Engineering, Stanford University \\
\email{christopher\_strong@berkeley.edu} 
\and
Department of Aeronautics and Astronautics, Stanford University \\ 
\email{\{smkatz, acorso, mykel\}@stanford.edu}}
\maketitle              
%

\setcounter{footnote}{0}

\begin{abstract}
Deep neural networks often lack the safety and robustness guarantees needed to be deployed in safety critical systems. Formal verification techniques can be used to prove input-output safety properties of networks, but when properties are difficult to specify, we rely on the solution to various optimization problems. 
In this work, we present an algorithm called \textsc{ZoPE} that solves optimization problems over the output of feedforward ReLU networks with low-dimensional inputs. The algorithm eagerly splits the input space, bounding the objective using zonotope propagation at each step, and improves computational efficiency compared to existing mixed-integer programming approaches. We demonstrate how to formulate and solve three types of optimization problems: (i) minimization of any convex function over the output space, 
(ii) minimization of a convex function over the output of two networks in series with an adversarial perturbation in the layer between them,
and (iii) maximization of the difference in output between two networks. Using \textsc{ZoPE}, we observe a $25\times$ speedup on property $1$ of the ACAS Xu neural network verification benchmark compared to several state-of-the-art verifiers, and an $85\times$ speedup on a set of linear optimization problems compared to a mixed-integer programming baseline. We demonstrate the versatility of the optimizer in analyzing networks by projecting onto the range of a generative adversarial network and visualizing the differences between a compressed and uncompressed network.


\end{abstract}

\section{Introduction}

\label{sec:related work}
The incorporation of deep neural networks (DNNs) into safety critical systems is limited by our ability to provide guarantees on their behavior \citep{BoDeDwFiFlGoJaMoMuZhZhZhZi16,Julian2019jgcd}. Neural network verification tools aim to provide these guarantees by proving whether a network satisfies a given input-output property \citep{liu2019algorithms}. When input-output relationships are difficult to specify, analyzing a system may require the solution to an optimization problem \citep{katz2021verification}. 

In this paper, we focus on solving optimization problems involving feedforward ReLU networks with low-dimensional inputs. Neural networks that control dynamical systems from state estimates often have low input dimension.
For example, the ACAS Xu networks for aircraft collision avoidance have a five-dimensional input \citep{Julian2019jgcd}. Additionally, semantic perturbations to high dimensional spaces can be analyzed through low dimensional networks \citep{katz2021verification}.
When the input space is low-dimensional, it can more easily be decomposed into smaller regions, each  defining a simpler optimization problem. We leverage this insight by rapidly dividing the input space into smaller regions that can be more tightly approximated, realizing a significant performance gain and finding the optimal value to a desired tolerance.

We consider the following three optimization problems, each of which is motivated by an application related to verifying the behavior of safety critical systems:
\begin{itemize}
    \item Minimizing a convex function of the output of a network. This problem can be used to reason about the actions of a control network \citep{julian2020validation,katz2021generating}. It can also be used to evaluate a generative adversarial network (GAN), which is a network architecture often used to model high-dimensional data distributions, by calculating the recall metric \citep{kynkaanniemi2019improved,katz2021verification}.
    \item Minimizing a convex function of the output of two networks in series subject to an adversarial attack at the output of the first network. This problem can be used to consider adversarial attacks on the input of a network when the input space is itself modeled by another network \citep{mirman2020robustness}.
    \item Maximizing the difference between the outputs of two networks given the same input. This problem can be used to compare a compressed and uncompressed network.
\end{itemize}
Minimizing a convex function of the output can be used to solve many neural network verification problems \citep{liu2019algorithms,strong2020global}. The other two problems have received less attention in the literature.

 In this work, we propose the \textbf{Zo}notope \textbf{P}ropagation with \textbf{E}agerness (\textsc{ZoPE}) optimizer, which solves these optimization problems to a desired tolerance by (i) eagerly breaking down the problem by splitting the input region, and (ii) relying on zonotope propagation to reason about the output reachable set
 from each input region. We consider a more “eager” solver to be one which spends less time on its bounding functions before splitting.
We evaluate the optimizer through runtime comparisons and qualitative demonstrations. 
We solve four of the standard ACAS Xu neural network verification benchmarks, and compare to state-of-the-art neural network verification tools \textsc{ERAN} \citep{singh2019boosting}, \textsc{nnenum} \citep{bak2020improved}, and \textsc{Marabou} \citep{katz2019marabou}. On property 1, which can be solved as a linear optimization problem over the output of the network, we observe a speedup of over $25 \times$ compared to the next best tool. We also evaluate the runtime of \textsc{ZoPE} on a batch of linear optimization problems from \citet{katz2021verification} and compare against a
baseline that mirrors \textsc{RefineZono}'s approach to verifying the ACAS Xu benchmark \citep{singh2019boosting}. We observe a speedup of $85\times$. Lastly, we demonstrate how \textsc{ZoPE} can be used as a tool to evaluate a generative adversarial network (GAN) and how it can be used to compare compressed to non-compressed networks.

There have been numerous recent works in the field of neural network verification.
These approaches often focus on networks with piecewise linear activation functions, such as the rectified linear unit (ReLU), and frequently take the form of a branch and bound search \citep{bunel2020branch}. Our optimizer does the same. Many break the verification problem into subproblems by case-splitting on the activation function or dividing the input domain \citep{ehlers2017formal,katz2017reluplex,katz2019marabou,bak2020improved,wang2018formal,DBLP:journals/corr/abs-2004-08440}. A survey by \citet{liu2019algorithms} compares these verification algorithms.

Many neural network verification tools can be extended to solve optimization problems \citep{tjeng2017evaluating,strong2020global}. Inspired by this idea, the proposed optimizer uses components from several verifiers --- it eagerly splits the input domain like \textsc{ReluVal} \citep{wang2018formal}, propagates zonotopes like \textsc{DeepZ} \citep{singh2018fast}, combines zonotope propagation with input splitting like \textsc{RefineZono} \citep{singh2019abstract}, and can optimize functions on the output like \textsc{MIPVerify} \cite{tjeng2017evaluating}. The pieces we drew from these different approaches were chosen in order to eagerly break down the input space while still limiting the overapproximation at each step. We expected rapidly splitting would have an advantage on networks with low-dimensional inputs that hadn't been fully explored by existing optimizers.
\\\\
This paper contains the following contributions:
\begin{itemize}
    \item A unified optimizer for three global optimization problems over low input dimension ReLU networks. These problems are of interest for verifying safety critical systems.
    \item A comparison of this new optimizer to existing verifiers and optimizers demonstrating a significant improvement against the state of the art when optimizing affine functions.
    \item Demonstrations of optimization problems which project onto the range of a network and find the maximum difference between two networks.
\end{itemize}

\section{Background}
\label{sec:background}
In this section we introduce notation, discuss the standard neural network verification problem, and compare it to the optimization problems that we focus on.
We view a network $f$ as representing a function
\[  
    f: \mathbb{R}^{n_{\text{in}}} \to \mathbb{R}^{n_{\text{out}}}
\]
We will only consider feedforward ReLU networks.
\\\\
\xhdr{Geometric objects and operations}
\label{subsec:geometric objects}
We will make use of several geometric objects. The first is a hyperrectangle, the generalization of a rectangle to $n$-dimensional space, which is defined by a center $\vec{c} \in \mathbb{R}^n$ and a radius $\vec{r} \in \mathbb{R}^n$ such that
\[
    H = \{\vec{x} \in \mathbb{R}^n \mid \vec{c} - \vec{r} \preceq \vec{x} \preceq \vec{c} + \vec{r} \}
\]
where $\preceq$ is the elementwise $\le$ between two vectors.

Hyperrectangles are a special case of a more general class of geometric objects called zonotopes, which can be defined as an affine transform of the unit hypercube. A zonotope $Z$ can be represented using matrix $\mat{G} \in \mathbb{R}^{n \times m}$ whose columns are referred to as generators, and a vector $\vec{c} \in \mathbb{R}^n$ which is the center of the zonotope as
\[
    Z = \{\vec{y} \in \mathbb{R}^n \mid \vec{y} = \mat{G} \vec{x} + \vec{c}, -1 \le x_i \le 1 \; \forall i = 1, \hdots, m\}
\]
Zonotopes are a subset of polytopes, and have symmetry about their center.
Optimizing a linear function over a hyperrectangle or a zonotope can be done analytically instead of by solving a linear program \citep{fujishige2005submodular,kitahara2019simple}.

We will also use the Minkowski sum between two sets $X$ and $Y$ defined as
\[
    X \oplus Y = \{\vec{x} + \vec{y} \mid \vec{x} \in X, \vec{y} \in Y\}
\]
This can be visualized as padding one set with the other.
\\\\
\xhdr{Zonotope Propagation}
A vital component of our approach will be finding an overapproximation of the output reachable set for a given input region. There are a variety of techniques 
%
to find symbolic or concrete descriptions of such a set \citep{wang2018formal,singh2018fast,liu2019algorithms}. One approach, used in the neural network verification tool \textsc{DeepZ} \citep{singh2018fast}, propagates zonotopes through a network layer by layer. After each layer the respective zonotope is an overapproximation of the reachable set for that layer. The new zonotope is formed elementwise, with overapproximation introduced for any dimension in the input zonotope that can be both negative and positive. For dimensions where this is true, an additional generator is introduced into the zonotope. The cost of computing this overapproximation is linear in the number of existing generators. We refer readers to the original paper, in particular Theorem 3.1, for details on this procedure \citep{singh2018fast}. We will make use of this algorithm in our optimizer, although in principal other overapproximate output reachable sets could be used. Exploring these alternatives is a promising direction for future work.


\section{Optimization Problems}
\label{sec:optimization problems}
The field of neural network verification has focused on checking input-output properties with yes or no answers. Formally, for input sets $\mathcal{X}$ and $\mathcal{Y}$ a neural network verification tool tells us whether the property
\begin{equation}
\label{eqn:verification problem}
    \vec{x} \in \mathcal{X} \implies \vec{y} \in \mathcal{Y}
\end{equation}
holds \citep{liu2019algorithms}. Recent work has explored extending these tools to solve optimization problems \citep{strong2020global}. In this work, we would like to address several optimization problems involving neural networks. In each problem we will only consider optimizing over hyperrectangular or zonotopic input sets.
\\\\
\xhdr{Minimizing a convex function on the range of a network} Our first problem of interest is to minimize a convex function on the output of a network. We can write this problem as
\begin{equation}
 \label{eqn:output-opt-convex}
    \begin{aligned}
        & \underset{\vec{x}}{\text{minimize}} && g(f(\vec{x})) \\
        & \text{subject to} 
        && \vec{x} \in \mathcal{X}
    \end{aligned}
\end{equation} 
where $g$ is a convex function. This can be used to solve a variety of neural network verification problems as defined in \cref{eqn:verification problem}. We can view the problem of projecting onto the range of a network as a special case with
\begin{equation} \label{eqn:projection}
    g(f(\vec{x})) = \norm{f(\vec{x}) - \vec{y}_0}
\end{equation}
An example use case is when $f$ is a generative adversarial network (GAN). By solving this optimization problem we can find the closest possible generated image to a ground truth image. 
\\\\
\xhdr{Noise buffer} We would like to optimize over the output of two networks in series with an adversarial perturbation applied between the two networks. This can be formulated as
 \begin{equation}
 \label{eqn:output-opt-with-noise-equation}
    \begin{aligned}
        & \underset{\vec{x}, \vec{z}}{\text{minimize}} && g(f_2(f_1(\vec{x}) + \vec{z}))) \\
        & \text{subject to} 
        && \vec{x} \in \mathcal{X} \\ 
        &&& \vec{z} \in Z
    \end{aligned}
\end{equation} 
where $Z$ is a zonotope of allowed perturbations and $f_1$ and $f_2$ are our two networks in series. The addition of $\vec{z}$ from the set $Z$ can be viewed as padding the output manifold of the first network. 
We will limit $g$ to be convex in this work. For an example of its use, consider if $f_1$ is a generative model and $f_2$ is a control network. By solving this optimization problem, we can evaluate the behavior of the controller with inputs defined by the generative model and subject to adversarial perturbations. Of note, this noise buffer optimization problem could also be put into the form of the first optimization problem in \cref{eqn:output-opt-convex} by considering an augmented input space that parameterizes the noise, then connecting those extra inputs to the intermediate layer with skip connections or a larger network. However, this could substantially increase the input dimension, so we focus on the framing of the problem given in \cref{eqn:output-opt-with-noise-equation} and leave a comparison with the alternative framing for future work.
\\\\
\xhdr{Network difference} A third optimization problem of interest is to determine how different the output of two networks can be if they take in the same input. We can write this as
 \begin{equation}
 \label{eqn:max-difference-equation}
    \begin{aligned}
        & \underset{\vec{x}}{\text{maximize}} && \norm{f_1(\vec{x}) - f_2(\vec{x})}_p \\
        & \text{subject to} 
        && \vec{x} \in \mathcal{X}
    \end{aligned}
\end{equation} 
for $\ell_p$ norm with $p \ge 1$. For an example of its use, consider if $f_1$ is a large network and $f_2$ is a smaller ``compressed'' network that attempts to mimic the behavior of $f_1$. By solving this optimization problem, we can evaluate how closely $f_1$ and $f_2$ will match. The non-convexity of this problem comes both from the network's non-convexity and from the fact that we would like to maximize rather than minimize a convex function.


\section{Approach}
\label{sec:approach}
Our proposed approach takes the form of a branch and bound search for the optimum value. The components within this branch and bound search will vary between optimization problems but share some common elements, including input splitting and zonotope propagation. Below we first sketch the general branch and bound algorithm and then discuss how it can be applied to each of the optimization problems of interest.

\subsection{Optimization With Branch and Bound}
 Branch and bound is an approach to optimization which repeatedly breaks down a problem into smaller sub-problems, bounding the optimal value of each sub-problem as it goes, and using those bounds to prune regions of the search space \citep{lawler1966branch,kochenderfer2019algorithms}. Suppose we would like to minimize an objective over some region. The branch and bound algorithm requires three functions: (i) \Call{Split}{}, (ii) \Call{UpperBound}{}, and (iii) \Call{LowerBound}{}. The function \Call{Split}{} splits a problem into multiple subproblems, \Call{LowerBound}{} finds a lower bound on the optimal value for a sub-problem, and \Call{upperBound} finds an upper bound on the optimal value for a sub-problem. The algorithm maintains a priority queue of subproblems ordered by their associated lower bound on the objective from \Call{LowerBound}{}, with highest priority given to the subproblem with the lowest lower bound. Some or all subproblems will also have associated upper bounds on their optimal value from \Call{UpperBound}{}. At each step, the subproblem with lowest lower bound is removed from the queue and split. Each new subproblem then has its lower bound evaluated and is added back onto the queue. The new subproblems may have an upper bound on their minimum objective evaluated as well, and those that don't inherit the upper bound of their parent subproblem.

The \emph{optimality gap} at any point is given by the difference between the lowest lower bound and the lowest upper bound across the 
open subproblems (those in the priority queue). If the optimality gap ever falls below a tolerance $\epsilon \ge 0$, the algorithm can return with a value within $\epsilon$ of the global optimum. The subproblems with lower bound greater than the lowest upper bound are effectively pruned, as they will never be revisited in the search for the optimum. If we would like to maximize instead of minimize an objective, we can reframe the problem as minimizing the negative of the original objective. Many neural network verification tools can be viewed as performing a branch and bound search for violations of a property \citep{bunel2020branch}.

In our case, the problem will correspond to an input set $\mathcal{X}$ that we would like to optimize over, and the subproblems will be regions from this original set. In this work we will only consider zonotope input sets, which includes hyperrectangles.
In order to solve the optimization problems described in \cref{sec:optimization problems} with the generic branch and bound algorithm, we will describe how to implement the three functions required: (i) \Call{Split}{}, (ii) \Call{UpperBound}{}, and (iii) \Call{LowerBound}{}.

\subsection{Split, UpperBound, LowerBound}
\label{subsec:each problem branch and bound}
We will start by addressing \Call{Split}{}, which will be common to each of the problems we would like to solve. For a zonotope input set $Z_{\text{in}} \subseteq \mathbb{R}^{n_{\text{in}}}$ defined by $n_{\text{gen}}$ generators $\mat{G} \in \mathbb{R}^{n_{\text{in}} \times n_{\text{gen}}}$ and center $\vec{c} \in \mathbb{R}^{n_{\text{in}}}$, we choose to split along the generator with largest $\ell_2$ norm using Proposition 3 from the work of \citet{althoff2008reachability}. This approach splits a zonotope into two zonotopes, but these zonotopes may have a non-empty intersection. Their union will be guaranteed to contain the original zonotope.

For a hyperrectangular input set, we choose the dimension with largest radius and split the hyperrectangle halfway along that dimension into two hyperrectangles. The interiors of the hyperrectangles will have an empty intersection. We experimented with a simple gradient based splitting heuristic but did not see an improvement to the performance. This may have been the result of the particular geometry of these networks. The computation required for the zonotope propagation at each step depends on the number of network activation regions, which are sets where the activation pattern of the network is constant, that overlap with the current input region.
As a result, we conjecture that a splitting strategy which aims to mold the subregions to match the geometric structure of the activation regions may be beneficial.
Other gradient or duality based input splitting heuristics from neural network verification tools may lead to better splits and should be explored in the future \citep{wang2018formal,rubies2019fast}.
Since we rely on splitting the input space, we expect our approach to scale poorly to high dimensions.


The approach to \Call{UpperBound}{} will also be similar across our problems. For the upper bound on the optimization problem over a region, we will evaluate the objective for a single point in the region. As an achievable objective, this will always upperbound the minimum achievable objective. We choose to evaluate the center of our input region. We experimented with a first order method to choose the point to evaluate but found limited benefit, and as a result chose to keep the heuristic of using the center point for simplicity. The optimality gap depends on two factors: the value of the achievable objective and the size of the input region. The overapproximation from propagating the input region is often more substantial, so choosing a better achievable objective does little to improve runtime.
As a result, even with a better heuristic there is a limit to the performance gains from the \Call{LowerBound}{} function.  
Many adversarial attacks could be repurposed to perform some local exploration for this step \citep{yuan2019adversarial}, and the tradeoff between the runtime of the \Call{UpperBound}{} function and the ability to reduce the optimality gap sooner could be explored. For the noise buffer problem, to find an upperbound we hold the input to the first network constant at the cell's center, leading to an output $\vec{y}_1$ from the first network. To account for points in the buffered region, we then optimize our objective over the second network with input given by the padded region $\{\vec{y}_1\} \oplus Z$.

Next, we will focus on \Call{LowerBound}{} for each of the optimization problems, which differs depending on the problem type. This function must map from a zonotopic or hyperrectangular input region $\mathcal{X}$ to a lower bound on the objective value.
\\\\
\xhdr{Minimizing a convex function on the range of a network} To lower bound a convex function over the output, we first propagate the input set $\mathcal{X}$ to a zonotopic output set $Z_{\text{out}}$ with generator $\mat{G}_{\text{out}} \in \mathbb{R}^{n_{\text{out}} \times n_{\text{gen}}}$ and center $\vec{c}_{\text{out}} \in \mathbb{R}^{n_{\text{out}}}$ which overapproximates the true output reachable set for this region. We then solve the convex program
 \begin{equation}
 \label{eqn:convex-program-on-zonotope}
    \begin{aligned}
        & \underset{\vec{z}}{\text{minimize}} && g(\vec{z}) \\
        & \text{subject to} 
        && \vec{z} \in Z_{\text{out}}
    \end{aligned}
\end{equation}
The constraint $\vec{z} \in Z_{\text{out}}$ is a set of linear constraints which can be written by introducing variables $\vec{x} \in \mathbb{R}^{n_{\text{gen}}}$ to get
 \begin{equation}
 \label{eqn:convex-program-on-zonotope-reframed}
    \begin{aligned}
        & \underset{\vec{z}, \vec{x}}{\text{minimize}} && g(\vec{z}) \\
        & \text{subject to} 
        && -1 \le x_i \le 1 \quad i = 1, \hdots, n_{\text{gen}} \\ 
        &&& \vec{z} = \mat{G}_{\text{out}}\vec{x} + \vec{c}_{\text{out}}
    \end{aligned}
\end{equation}
We will return the optimal value $p^*$ of this convex program as the lower bound.

If $g$ is an affine function $g(\vec{y}) = \vec{a}^\top \vec{y} + b$, then the solution is analytic and is given by 
\begin{equation}
    p^* = \vec{c}_{\text{out}}^\top \vec{a} + \norm{\mat{G}_{\text{out}}^\top \vec{a}}_1 + b
\end{equation}
where $\mat{G}$ is the generator matrix for the zonotope and $\vec{c}$ is the center of the zonotope \citep{althoff2016combining}. Computing this expression will typically be much faster than solving a convex program, giving a large speedup when optimizing an affine function. 


Additionally, checking whether the output of a network is always contained within a polytope $\mathcal{P} = \{\vec{x} \mid \mat{A} \vec{x} \le \vec{b}, \vec{A} \in \mathbb{R}^{n \times m}, \vec{b} \in \mathbb{R}^{n} \}$ can be accomplished by maximizing the maximum violation of the polytope's constraints. We will denote the $i$th row of $\mat{A}$ as $\vec{a}_i^\top$. This problem could either be solved with the above framework through $n$ separate queries with the negative violation of the $i$th constraint as the objective $g(\vec{y}) = -\vec{a}_i^\top \vec{y} - b_i$, or through a single query with 
$g(\vec{y}) = -\max_i (\vec{a}_i^\top \vec{y} - b_i)$
This objective is the negative of a pointwise maximum of affine functions, so is concave. Fortunately, although $g$ is concave, minimizing $g$ over a zonotope can be accomplished with one linear optimization per row of $\mat{A}$, each of which is analytical. As a result, checking whether the output of a network is always contained within a polytope $\mathcal{P}$ can be performed through $n$ separate queries which solve a single linear optimization at each step, or through a single query which solves $n$ linear optimizations at each step.

Lastly, if we are projecting onto the range of a network with $g(\vec{y}) = \norm{\vec{y} - \vec{y}_0}$, the choice of norm will affect the complexity of the optimization problem over a zonotope. For example, with $\ell_1$ or $\ell_\infty$ norms this can be formulated as a linear program, while for the $\ell_2$ norm it will be a quadratic program. Future work could explore using faster projection algorithms instead of solving a convex program at each step which may yield significant speedups.
\\\\
\xhdr{Noise buffer}
We would like to optimize a function over two networks in series with a buffer of allowed perturbations $Z$ after the first layer. This is equivalent to taking the Minkowski sum of the output manifold of the first network and the buffer. We would like to find a lower bound on the objective that will approach the true objective as the input cell grows smaller. We first propagate the cell through the first network to get a zonotope $Z_{1_{\text{out}}}$ which overapproximates the reachable set. We then take the Minkowski sum of this zonotope with our buffer to get
\[
Z_{\text{buffered}} = Z_{1_{\text{out}}} \oplus Z = \{\vec{z}_{1_{\text{out}}} + \vec{z} \mid \vec{z}_{1_{\text{out}}} \in Z_{1_{\text{out}}}, \vec{z} \in Z\}
\]
Since zonotopes are closed under Minkowski sums, the resulting object will still be a zonotope \citep{althoff2015computing}.

Our problem now becomes trying to lower bound our function $g$ on this buffered set. As our input cell becomes small, $Z_{1_{\text{out}}}$ does as well, and $Z_{\text{buffered}}$ approaches the size of the buffer. Since the buffered zonotope will not become arbitrarily small, if we were to just propagate $Z_{\text{buffered}}$ through the second network, we would incur some steady state error in our lower bound. To avoid this overapproximation, we can solve the optimization problem from the buffered zonotope to the output exactly. If the dimension of the intermediate space is low, we could apply the algorithm we have already given for optimizing convex functions over a single network. If the dimension is high, we can use another optimization strategy such as encoding the second network using mixed-integer constraints as done by \textsc{NSVerify}, \textsc{MIPVerify}, and \textsc{ERAN} \citep{lomuscio2017approach,tjeng2017evaluating,singh2019boosting}, then adding the objective and solving the resulting optimization problem with an off-the-shelf MIP solver such as Gurobi or GLPK. Since this approach nests another full optimization problem over the second half of the network within each step of the original branch and bound, we expect the runtime to scale poorly as the size of the perturbation set $Z$ and the complexity of the second network grow, which may limit the use of the proposed approach for this type of analysis.

In summary, to get a lower bound we (i) overapproximate the set passing through the first network, then (ii) solve the resulting optimization problem over the second network with input set given by a buffered zonotope.
\\\\
\xhdr{Network difference} Our goal is to find the maximum difference in the output of two networks over an input region. Since we are maximizing a function, we are interested in finding an upper bound on the objective over our input cell. We start by propagating the input cell through the first network to get $Z_{1_{\text{out}}}$ and the second network to get $Z_{2_{\text{out}}}$. We can then tightly overapproximate each of these zonotopes as hyperrectangles $H_1$ and $H_2$ by finding their maximum and minimum value in each elementary direction. Each of these operations can be performed analytically. Once we have these two hyperrectangular overapproximations, we are interested in solving
\begin{equation}
 \label{eqn:hyperrectangular-distance}
    \begin{aligned}
        & \underset{\vec{h}_1, \vec{h}_2}{\text{maximize}} && \norm{\vec{h}_1 - \vec{h}_2}_p \\
        & \text{subject to} 
        && \vec{h}_1 \in H_1 \\
        &&& \vec{h}_2 \in H_2
    \end{aligned}
\end{equation}
whose optimal value will upper bound the true maximum distance in this region. Let $\vec{c}_1$ and $\vec{c}_2$ be the centers of $H_1$ and $H_2$ and $\vec{r}_1$ and $\vec{r}_2$ be the radius of $H_1$ and $H_2$ in each elementary direction. An analytical solution to this optimization problem is given by 
\begin{align*}
    \vec{h}_1^* &= \vec{c}_1 + \text{sign}(\vec{c}_1 - \vec{c}_2) \odot \vec{r}_1 \\
    \vec{h}_2^* &= \vec{c}_2 + \text{sign}(\vec{c}_2 - \vec{c}_1) \odot \vec{r}_2 \\
    d^* &= \norm{\vec{h}_1^* - \vec{h}_2^*}_p
\end{align*}
where $\odot$ represents elementwise multiplication and $d^*$ is the optimal value. See Appendix A.1 for a derivation of this analytical solution.
Returning $d^*$ as defined above will upper bound the objective function.



\subsection{Implementation}
\label{subsec:implementation}
Each of the approaches described in \cref{subsec:each problem branch and bound} were implemented in a Julia package.\footnote{Source is at 
\ifanonymize
 [[link omitted for double blind review]]. 
\else
 \url{https://github.com/sisl/NeuralPriorityOptimizer.jl}.
\fi
}
This repository also has code to reproduce the benchmarks on our optimizer in \cref{sec:experimental results}.
The zonotope propagation and zonotope splitting is performed with the LazySets library.\footnote{Source is at \url{https://github.com/JuliaReach/LazySets.jl}.} 
For solving linear and mixed-integer linear programs we use Gurobi and for solving other convex programs we use Mosek, both of which have a free academic license.\footnote{Available at \url{https://www.gurobi.com} and \url{https://www.mosek.com}.} The implementation is modular and is intended to be easily extended to solve other optimization problems. 

\section{Experimental Results}
\label{sec:experimental results}
We apply \textsc{ZoPE} to a variety of problems, first comparing its runtime to existing solvers on the ACAS Xu benchmark and linear optimization problems. We then showcase how it can be used to solve problems with more complex objectives. 
In several of these experiments we use a conditional GAN trained to represent images from a wing-mounted camera on a taxiing aircraft. The conditional GAN has four inputs, two of which are the crosstrack position and heading while the other two are latent inputs. We also use a state estimation network which takes as input a 128-dimensional image of the taxiway and outputs the state of the aircraft. The GAN and state estimation network can be combined in series. All timing is done on a single core of an Intel Xeon 2.20GHz CPU and with an optimality gap of \num{1e-4} unless otherwise specified. All queries use hyperrectangular input sets; in future work it would be valuable to explore the runtime consequences when splitting non-hyperrectangular input zonotopes as well.

\subsection{ACAS Xu Benchmark}
\label{subsec:acas xu}
The ACAS Xu neural network verification benchmark contains a set of properties on networks trained to compress the ACAS Xu collision avoidance system and is often used to benchmark verification tools \citep{katz2017reluplex,Julian2019jgcd}.
We will consider properties $1$, $2$, $3$, and $4$ introduced by \citet{katz2017reluplex}. We compare to the neural network verification tools \textsc{Marabou} \citep{katz2019marabou}, \textsc{nnenum} \citep{bak2020improved}, and \textsc{ERAN} \citep{singh2018fast,singh2019abstract,singh2019beyond,singh2019boosting}. See Appendix A.2 for details on how each solver was configured.
Property 1 can be evaluated by maximizing a linear function, while properties $2$, $3$, and $4$ can be evaluated by minimizing the convex indicator function to the output polytope associated with the property or by minimizing the distance to the output polytope associated with the property. Viewed in another way, property 1 can be solved by asking the question ``Is the network always contained in a polytope?'' while property 2 can be solved by asking the question ``Does the network ever reach a polytope?'' For property $1$ each step is analytical, while for properties $2$, $3$, and $4$ at each step we apply a quick approximate check for intersection, and if it is indeterminate we solve a linear program. Each verification tool was run on a single core.

\Cref{fig:acas} shows the performance of the optimizer on four ACAS properties. 
\textsc{ZoPE} achieves a speedup of about $25 \times$ on property 1. We remain competitive with the other tools on properties 2, 3, and 4, where we may need to solve a linear program at each step.

\begin{figure*}[t]
    \centering
    \input{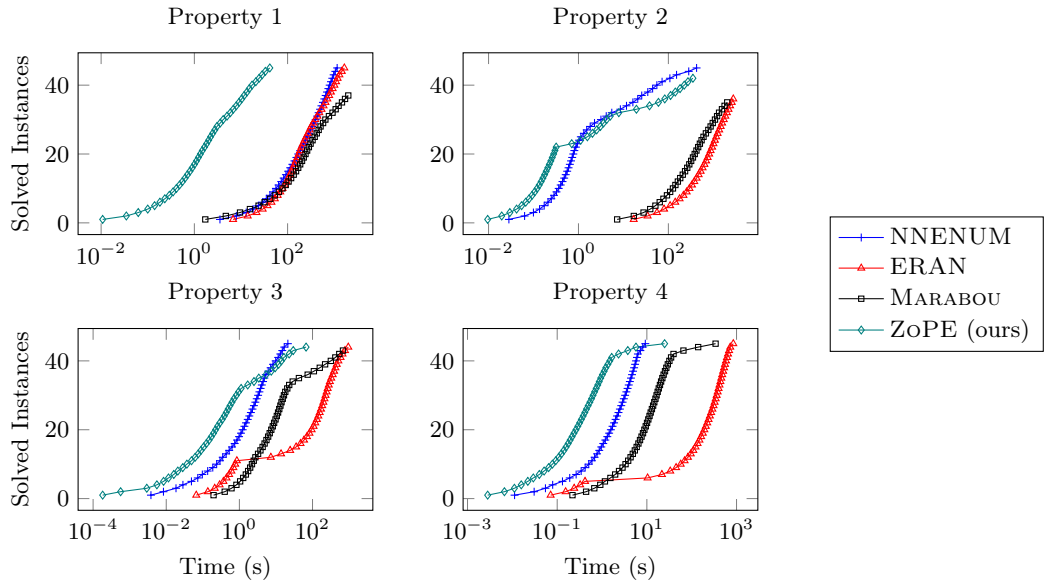}
    \caption{Comparison of Solvers on ACASXu Properties 1, 2, 3, and 4 with a 300 second timeout. \label{fig:acas}}
\end{figure*}

\subsection{Optimizing Convex Functions} 
\label{subsec:optimizing convex functions}
We first evaluate \textsc{ZoPE} maximizing a linear objective. We run queries on a network composed of the conditional GAN concatenated with the image-based control network. This combined network was introduced by \citet{katz2021verification} and has an input of two states and two latent dimensions. The objective function corresponds to the control effort. The baseline we compare against divides the state dimensions into hyperrectangular cells, propagates a zonotope through each cell with \textsc{DeepZ}'s approach, then uses the resulting bounds to formulate a mixed-integer program and find the optimum for that cell. Since we run these queries sequentially, each mixed-integer program also has a constraint that the objective should be larger than the best seen so far.
The strategy of interleaving splitting and MIP calls mirrors \textsc{RefineZono}'s approach to verifying the ACAS Xu networks \citep{singh2019boosting}. 
\Cref{tab:optimization performance} shows more than an $85 \times$ speedup of our approach over the baseline.
The efficiency of \textsc{ZoPE} relies heavily on the computational cost of finding bounds for the objective over a zonotope. As a result, like with ACAS property $1$ we see substantially better performance than existing tools when optimizing an objective with only analytical operations at each step.


\newcolumntype{x}[1]{%
>{\raggedleft\hspace{0pt}}p{#1}}%

\begin{table}[t!]
    \centering
    \caption{Performance on linear optimization problems. 25 queries in different regions of the input space are run on a single network. The network was introduced in \citet{katz2021verification} and consists of a conditional GAN concatenated with an image-based controller. The performance of the MIP approach with a variety of discretizations of the state space is shown. For example, MIP $3 \times 3$ corresponds to an optimizer which for each query (i) discretizes the input space into a $3 \times 3$ grid, then (ii) for each cell in the grid finds bounds on each node using the approach of \textsc{DeepZ}, and (iii) solves the resulting MIP using Gurobi. \label{tab:optimization performance}}
    \vspace{1ex}
    \begin{tabular}{@{}l@{\hskip 0.2in}r@{}} 
        \toprule
        \textbf{Approach} & \textbf{Total Time (s)} \\
        \midrule
        MIP $3 \times 3$ & $3728$ \\
        MIP $5 \times 5$ & $1171$ \\
        MIP $10 \times 10$ & $1610$ \\
        MIP $15 \times 15$ & $2473$ \\
        ZoPE (ours) & $\mathbf{13.5}$ \\
        \bottomrule
    \end{tabular}
\end{table}



Next, we demonstrate using the proposed optimizer to project an image onto the output manifold of a conditional GAN. The GAN has a finite, convex support for its latent variables. This allows us to project onto the range of the network, under some $\ell_p$ norm, by minimizing the convex objective function in \cref{eqn:projection}. \Cref{fig:close_ims} shows several images and their corresponding closest generated images from the GAN. The visual similarity between the two rows gives some evidence that the GAN is capturing the desired images in its output manifold. However, we still see some slight differences between the images. The degree of these differences can be used to measure how closely the GAN captures each training datapoint, giving a recall metric to evaluate a GAN and inform hyperparameter choice, as was done in \citet{katz2021verification}. Note that this analysis, and the sense of ``closeness'' in this context, depends on the norm used for the projection.
\begin{figure*}[t!]
    \centering
    \input{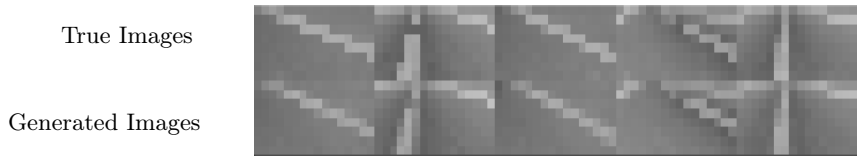}
    \caption{Closest generated images (bottom row) to a set of true images (top row) with distance measured by the $\ell_2$ norm. \label{fig:close_ims}}
\end{figure*}


\subsection{Maximum Distance Between Compressed and Original Networks}
By finding the maximum distance between the outputs of two networks as described in \cref{subsec:each problem branch and bound}, we can evaluate how well a compressed network mimics the behavior of an original uncompressed network. We validate this technique on a large conditional GAN, with two input states to be conditioned on, two latent dimensions, four layers with 256 ReLUs each, and a 128 dimensional output layer. The second ``compressed'' network has the same input and output spaces, but only two layers with 128 ReLUs each. We use a required optimality gap of \num{0.1}. The heatmap in \cref{fig:max_output} shows the maximum difference in the output of these networks across a slice of the state space. These maximum differences, or an approximation thereof, could be used to retrain the network in regions where the difference is large.

\begin{figure}[t!]
    \centering
    \input{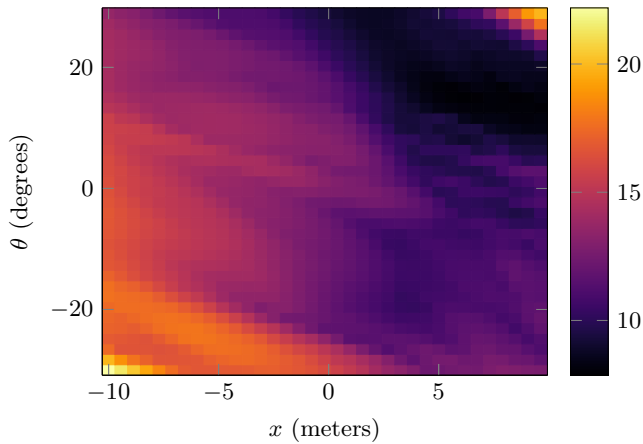}
    \caption{The maximum output distance in L1 norm of two networks over the state space.}
    \label{fig:max_output}
\end{figure}

\section{Conclusion}
\label{sec:conclusion}
In this work, we introduced an algorithm for solving a wide variety of optimization problems on feedforward ReLU networks with low input dimension. The algorithm relies on eagerly splitting the input space and making use of zonotope propagation through the network to bound the optimum at each step. 
We observe a speedup of $25 \times$ on property $1$ of the ACAS Xu benchmark compared to several existing verification tools, and $85 \times$ on a linear optimization benchmark compared to a mixed-integer programming baseline. We also demonstrate how the optimizer can be used to analyze how closely a GAN has learned to replicate its training data and how it can be used to compare a compressed and uncompressed network. The optimizer was implemented modularly and was made available as a Julia package at 
\ifanonymize
[[link omitted for double blind review]]
\else
\url{https://github.com/sisl/NeuralPriorityOptimizer.jl}
\fi 
so as to flexibly allow for a reader to explore solving other optimization problems. Any non-convex objective which can be optimized over a zonotope can readily be optimized in this framework, as was demonstrated in both our approach to check whether the output of a network is contained within a polytope and to maximize the distance between the output of two networks.

There are several major avenues for future work. The often prohibitive growth of the runtime with the input dimension, depth, and width of the network remains as a significant challenge for this and other exact optimizers. One direction of interest would be to develop more specialized lower bound functions for particular problems. For example, faster intersection or projection algorithms may be applied to some problems where our implementation solves a convex program at each step.
We could also incorporate and compare some of the optimizations that ERAN makes use of; for example, mixing mixed-integer program solves in with the splitting, 
tightening the propagated zonotopes, or propagating polytopes instead of zonotopes. Another would be to consider how to scale up to high-dimensional input spaces, and consider what a more eager splitting strategy looks like in those contexts. Lastly, we could find other optimization problems of interest over neural networks that could be solved with the same or a similar framework. 


\ifanonymize 

\else

\section*{Acknowledgments}


We would like to acknowledge support from Eric Luxenberg, Haoze Wu, Gagandeep Singh, Chelsea Sidrane, Joe Vincent, Changliu Liu, Tomer Arnon, and Katherine Strong.

Funding in support of this work is from DARPA under contract FA8750-18-C-009, the NASA University Leadership Initiative (grant \#80NSSC20M0163), and the National Science Foundation Graduate Research Fellowship under Grant No. DGE–1656518. Any opinions, findings, and conclusions or recommendations expressed in this material are those of the authors and do not necessarily reflect the views of DARPA, any NASA entity, or the National Science Foundation.

\fi

\typeout{}
\printbibliography
\newpage 

\appendix

\section{Appendix}
\subsection{Maximum Distance Between Points in Two Hyperrectangles}
\label{appendix:max distance proof}

We would like to derive an analytical solution for the maximum distance given by a $p$-norm with $p \ge 1$ between two hyperrectangles $H_1$ and $H_2$. We will let $\vec{c}_1$ and $\vec{c}_2$ be the centers of $H_1$ and $H_2$, and $\vec{r}_1$ and $\vec{r}_2$ be the radii of $H_1$ and $H_2$. The maximum distance can be found by solving the following optimization problem
\begin{equation*}
    \begin{aligned}
        & \underset{\vec{h}_1, \vec{h}_2}{\text{maximize}} && \norm{\vec{h}_1 - \vec{h}_2}_p \\
        & \text{subject to} 
        && \vec{h}_1 \in H_1 \\
        &&& \vec{h}_2 \in H_2
    \end{aligned}
\end{equation*}
The $p$-norm for finite $p$ is defined as 
 \[
    \norm{\vec{x}}_p = (\sum_{i=1}^n |(\vec{x})_i|^p)^{\frac{1}{p}}
 \]
We expand the objective of our maximization problem to be
\[
    (\sum_{i=1}^n (|(\vec{h}_1)_i - (\vec{h}_2)_i|^p))^{\frac{1}{p}}
\]
and since $x^\frac{1}{p}$ is monotonically increasing on the non-negative reals for $p \ge 1$, we can remove the power of $\frac{1}{p}$ giving us the equivalent problem
\begin{equation}
 \label{eqn:hyperrectangular-distance-reformulated}
    \begin{aligned}
        & \underset{\vec{h}_1, \vec{h}_2}{\text{maximize}} && \sum_{i=1}^n (|(\vec{h}_1)_i - (\vec{h}_2)_i|^p) \\
        & \text{subject to} 
        && \vec{h}_1 \in H_1 \\
        &&& \vec{h}_2 \in H_2
    \end{aligned}
\end{equation}
Now we see that the constraints $\vec{h}_1 \in H_1$ and $\vec{h}_2 \in H_2$ apply independent constraints to each dimension of $\vec{h}_1$ and $\vec{h}_2$. We also note that the objective can be decomposed coordinate-wise. As a result, in order to solve this optimization problem, we will need to solve $n$ optimization problems of the form
\begin{equation}
 \label{eqn:hyperrectangular-distance-reformulated-elementwise}
    \begin{aligned}
        & \underset{(\vec{h}_1)_i, (\vec{h}_2)_i}{\text{maximize}} && |(\vec{h}_1)_i - (\vec{h}_2)_i|^p \\
        & \text{subject to} 
        &&  (\vec{c}_1)_i - (\vec{r}_1)_i \le (\vec{h}_1)_i \le (\vec{c}_1)_i + (\vec{r}_1)_i\\
        &&&  (\vec{c}_2)_i - (\vec{r}_2)_i \le (\vec{h}_2)_i \le (\vec{c}_2)_i + (\vec{r}_2)_i
    \end{aligned}
\end{equation}
Since $x^p$ is monotonically increasing for $p \ge 1$ we can equivalently maximize $|(\vec{h}_1)_i - (\vec{h}_2)_i|$. We show an analytic form for the maximum by checking cases.
If $(\vec{c}_2)_i$ is larger than $(\vec{c}_1)_i$, the maximum will be found by pushing $(\vec{h}_2)_i$ to its upper bound and $(\vec{h}_1)_i$ to its lower bound. Conversely, if $(\vec{h}_1)_i$ is larger than $(\vec{h}_2)_i$, the maximum will be found by pushing $(\vec{h}_1)_i$ to its upper bound and $(\vec{h}_2)_i$ to its lower bound. If $(\vec{c}_1)_i$ is equal to $(\vec{c}_2)_i$, then we can arbitrarily choose one to push to its lower bound and the other to push to its upper bound --- we select $(\vec{h}_1)_i$ to go to its upper bound and $(\vec{h}_2)_i$ to go to its lower bound. As a result we have the optimal inputs
\begin{align*}
    (\vec{h}_1)_i^* &= (\vec{c}_1)_i + \text{sign}((\vec{c}_1)_i - (\vec{c}_1)_i) (\vec{r}_1)_i \\
    (\vec{h}_2)_i^* &= (\vec{c}_2)_i + \text{sign}((\vec{c}_2)_i - (\vec{c}_2)_i) (\vec{r}_2)_i
\end{align*}
where the sign function is given by
\[
    \text{sign}(x) = \begin{cases} 1.0 & x \ge 0 \\ -1.0 & x < 0 \end{cases}
\]
Then, backtracking to our original problem and vectorizing gives us the analytical solution to this optimization problem with optimal value $d^*$
\begin{align*}
    \vec{h}_1^* &= \vec{c}_1 + \text{sign}(\vec{c}_1 - \vec{c}_2) \odot \vec{r}_1 \\
    \vec{h}_2^* &= \vec{c}_2 + \text{sign}(\vec{c}_2 - \vec{c}_1) \odot \vec{r}_2 \\
    d^* &= \norm{\vec{h}_1^* - \vec{h}_2^*}_p
\end{align*}
where the sign function is applied elementwise. This completes our derivation of the analytical solution for the maximum distance between two points contained in two hyperrectangles.

\subsection{Verifier Configuration for the Collision Avoidance Benchmark}

This section describes how each verifier was configured for the collision avoidance benchmark discussed in section 5.1. \Cref{table:verifier configuration} summarizes the non-default parameters for each solver and the location where the parameter was set. Both \textsc{NNENUM} and \textsc{ERAN} by default make use of parallelization, and \textsc{Marabou} has a parallel mode of operation, but for this experiment we restrict all tools to a single core. We ran the experiments on a single core to try to separate the aspects of how each solver was parallelized from what we viewed as the core of its algorithmic approach. We expect \textsc{ZoPE} would parallelize well, especially on more challenging problems. The hyperparameters we ran for \textsc{ERAN} may be better suited for multiple cores than a single core, so further comparison could explore these in more depth. Additionally, the timing results from the Verification of Neural Networks 2020 competition\footnote{\url{https://sites.google.com/view/vnn20/vnncomp}} for several properties for \textsc{ERAN} were slower than we expected from the change in hardware and the restriction to a single core. Exploring the tool further, we observed that on several problem instances it would return back a failed status before reaching a timeout. On these same instances we saw that \textsc{ERAN} would find several inputs that were almost counter-examples, for example with a margin of \num{1e-6} from violating the property, flag these as potential counter-examples, then move on. It is possible that the root cause of the abnormalities we observed affected timing results. On problems where \textsc{ERAN} did return a status the results were consistent with the ground truth.   

The parameters were chosen based off of a mix of recommendations from developers on their best configuration for the collision avoidance benchmark or existing documented settings for this benchmark. For example, \textsc{ERAN}'s parameters were based off of the VNN20 competition as found at \url{https://github.com/GgnDpSngh/ERAN-VNN-COMP/blob/master/tf_verify/run_acasxu.sh}.  The code for for \textsc{Marabou},\footnote{\url{https://github.com/NeuralNetworkVerification/Marabou}} \textsc{NNENUM},\footnote{\url{https://github.com/stanleybak/nnenum}} \textsc{ERAN},\footnote{\url{https://github.com/eth-sri/eran}} and our optimizer \textsc{ZoPE}\footnote{
\ifanonymize
 [[link omitted for double blind review]]. 
\else
 \url{https://github.com/sisl/NeuralPriorityOptimizer.jl}
\fi
} is available for free online.

\begin{table}[htb]
    \centering
    \caption{Non-Default Verifier Parameters  \label{table:verifier configuration}}
    \begin{small}
    \begin{tabular}{@{}llrr@{}}
         \toprule
         \textbf{Solver} &  \textbf{Parameter} & \textbf{Value} & \textbf{Location} \\
         \midrule
         \textsc{Marabou} & & & \\
         \arrayrulecolor{black!30} \midrule
          & split-threshold & 1 & Command line argument \\
          & INTERVAL\_SPLITTING\_FREQUENCY & 1 & GlobalConfiguration.cpp file \\
         \arrayrulecolor{black}\midrule
         \textsc{NNENUM} & &  & \\
         \arrayrulecolor{black!30} \midrule 
          & Settings.NUM\_PROCESSES & 1 & acasxu\_all.py file\\ 
         \arrayrulecolor{black}\midrule
         \textsc{ERAN} & & & \\ 
         \arrayrulecolor{black!30} \midrule 
         & use\_parallel\_solve & True & \_\_main\_\_.py file\\ 
         & processes & 1 & \_\_main\_\_.py file \\ 
         & domain & deeppoly & Command line argument \\ 
         & complete & True & Command line argument \\ 
         & timeout\_milp & 10 & Command line argument \\ 
         & numproc & 1 & Command line argument \\ 
        \arrayrulecolor{black}\midrule
         \textsc{ZoPE} & & & \\ 
         \arrayrulecolor{black!30} \midrule 
         & stop\_gap & \num{1e-4} & acas\_example.jl \\
         & stop\_frequency & 1 & acas\_example.jl \\ 
         \arrayrulecolor{black}\bottomrule
    \end{tabular}
    \end{small}
\end{table}

\fi

\end{document}